\DeclareUnicodeCharacter{FF0C}{fi}
\documentclass{article}
\usepackage{spconf,amsmath,epsfig}
\usepackage{amssymb}
\usepackage{color}

\let\OLDthebibliography\thebibliography
\renewcommand\thebibliography[1]{
  \OLDthebibliography{#1}
  \setlength{\parskip}{0pt}
  \setlength{\itemsep}{0pt plus 0.3ex}
}
\begin{document}\sloppy

\def\x{{\mathbf x}}
\def\L{{\cal L}}

\title{Grad-CAM guided channel-spatial attention module \\for Fine-grained visual classification }
%
\name{Shuai Xu, Dongliang Chang, Jiyang Xie, and Zhanyu Ma*}
\address{}

\maketitle

\begin{abstract}
Fine-grained visual classification (FGVC) is becoming an important research field, due to its wide applications and the rapid development of computer vision  technologies. The current state-of-the-art (SOTA) methods in the FGVC usually employ attention mechanisms to first capture the semantic parts and then discover their subtle differences between distinct classes. The channel-spatial attention mechanisms, which focus on the discriminative channels and regions simultaneously, have significantly improved the classification performance. However, the existing attention modules are poorly guided since part-based detectors in the FGVC depend on the network learning ability without the supervision of part annotations. As obtaining such part annotations is labor-intensive, some visual localization and explanation methods, such as gradient-weighted class activation mapping (Grad-CAM), can be utilized for supervising the attention mechanism. We propose a Grad-CAM guided channel-spatial attention module for the FGVC, which employs the Grad-CAM to supervise and constrain the attention weights by generating the coarse localization maps. To demonstrate the effectiveness of the proposed method, we conduct comprehensive experiments on three popular FGVC datasets, including CUB-$200$-$2011$, Stanford Cars, and FGVC-Aircraft datasets. The proposed method outperforms the SOTA attention modules in the FGVC task. In addition, visualizations of feature maps also demonstrate the superiority of the proposed method against the SOTA approaches.
\end{abstract}
\section{Introduction}
\label{sec:intro}

Fine-grained visual classification (FGVC) aims to distinguish fine-grained classes under the same coarse class labels,~\emph{e.g.}, birds~\cite{birds}, airplanes~\cite{air}, cars~\cite{cars}， and flowers~\cite{flowers},~\emph{etc.} The main challenge of the FGVC task is the tiny inter-class difference along with significant intra-class diversity. For example, it is difficult to distinguish a redhead woodpecker from a pileated woodpecker and a downy woodpecker caused by highly similar sub-categories, but with the adjustment of poses, scales, and rotations, the redhead woodpecker can be photographed in a very different visual view. In order to generate discriminative features more precisely, we better have the ability to capture the key characteristics of the red head and ignore the background and other irrelevant regions, which is an obvious way for overcoming the challenge. 

\begin{figure}
	\centering 
	\includegraphics[width=1\linewidth]{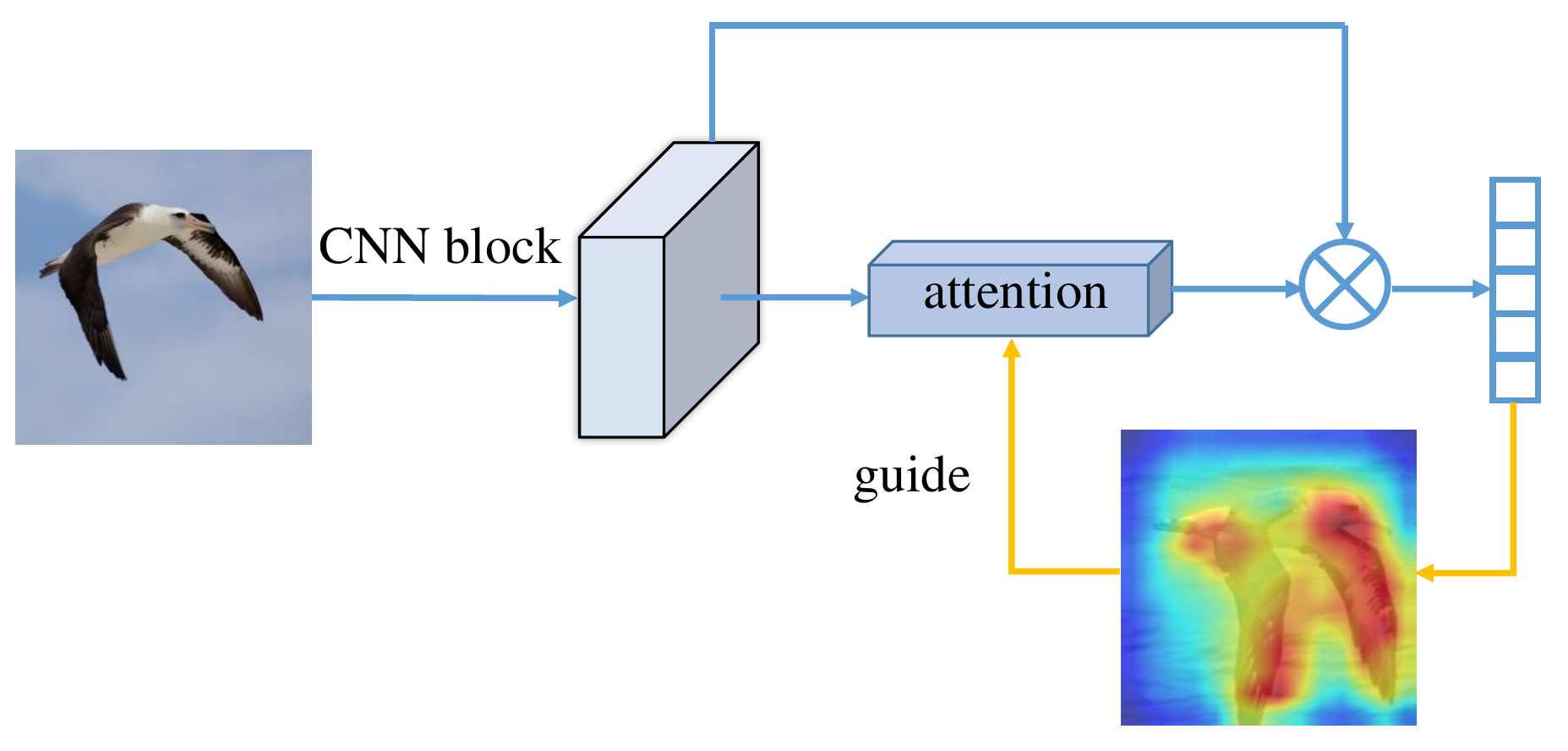}  
	\vspace{-6mm}
	\caption{Motivation of the Grad-CAM guided channel-spatial attention module. The blue part is the general pipeline of the previous attention mechanisms. The yellow line is our proposed supervision mechanism that the weights obtained by the gradient backpropagation in the Grad-CAM are used for the guidance of the attention weights, with which the attention mechanisms focus on parts that contribute significantly to classification.}\label{fig:attention}
	
\end{figure}

    The existing approaches can be roughly divided into two classes: ($1$) searching the informative regions that contribute the most to the classification task~\cite{anno,TASN,NTS} and ($2$) paying more attention to extract high-order features from the images~\cite{BP,BT,mamcloss,mc}. For the former one, previous approaches~\cite{anno} usually employed the prior location information such as part-level bounding boxes and segmented masks to generate the discriminative parts. Meanwhile, for the latter one, the powerful deep networks~\cite{BP,BT} were employed for feature extraction, and different loss functions~\cite{mamcloss,mc} were designed for constraining these networks to improve the discrimination of the extracted features. Recently, the attention mechanisms~\cite{SENet,MA-CNN,RA-CNN}, which only require image labels for training, have gradually replaced the   manual annotation methods, since part annotations are time-consuming and laborious that limits the flexibility and versatility of the real-world FGVC applications. Compared with the approaches that introduce complex structures, the attention-based methods add fewer learned parameters for model training, which efficiently reduces the computation cost.

The attention mechanisms fully simulate the observation habits of human eyes, which always concentrate on the most distinctive regions for observing. For example, we can easily pay attention to the head and the wings of a bird and ignore the other common regions to identify its species. Based on this motivation, many methods have been proposed by utilizing the attention mechanisms to detect the discriminative information from the images, including channel attention~\cite{TASN,SENet}, spatial attention~\cite{RA-CNN,PA-CNN}, and channel-spatial attention~\cite{CBAM}. Specifically, SENet~\cite{SENet} introduced ``squeeze-and-excitation'' (SE) blocks to adaptively recalibrate the feature maps in channel-wise by modeling the interactions between channels. The trilinear attention sampling network~\cite{TASN}  generated attention maps by integrating feature channels with their relationship matrix and highlighted the attended parts with high resolution.  The recurrent attention convolutional neural network (RA-CNN)~\cite{RA-CNN} introduced attention proposal network (APN) to capture the region relevance information based on the extracted features, and then amplified the attention region crops to make the network gradually focus on the key areas. The convolutional block attention mechanism (CBAM)~\cite{CBAM} is a channel-spatial attention method that utilizes both the channel-level and region-level information. It can effectively improve the characteristic expression ability of the networks. The existing methods~\cite{TASN,SENet,RA-CNN,CBAM} usually utilize different attention mechanisms to generally adjust the distributions of the attention weights for balancing the contributions of feature maps extracted from each 
part. Although these methods for obtaining the weights are different, they are all constructed based on the original feature maps only, without part information  supervision. Obviously, if the feature maps focus on the non-significant parts such as backgrounds and distractions, the attention mechanism is meaningless under the unsupervised conditions. 

To address this issue, we propose a weakly-supervised guideline for discriminative part mining and informative feature learning. It drives the networks to focus on the parts which have specific characteristic information, such as the head and the beak of a bird. In addition, as each channel of the feature maps can be also considered as a semantic part~\cite{MA-CNN}, supervision on discriminative parts can be transferred to that on channels. Gradient-weighted class activation mapping (Grad-CAM)~\cite{cam} is usually introduced to illustrate attentions of the networks with heat maps and visualize the attentions in each part by weighted averaging channels, which can be used for guiding the networks to focus on more efficient parts and discard the redundant information for the classification. In this paper, we introduce a channel-spatial attention mechanism with Grad-CAM guided, in which the channel weighted feature maps are pooled along with the channel dimensions and multiplied by the original feature maps to obtain the channel-spatial attention maps. Meanwhile, a novel Grad-CAM guided channel-spatial attention mechanism loss (GGAM-Loss) is applied for guiding the learning process of the channel weights and forcing the attention module to focus on the parts that contribute most to the classification. As shown in Figure \ref{fig:attention}, we employ the channel weights obtained from the gradient backpropagation in the Grad-CAM to constrain the channel weights of the forward propagation. 

Our contributions can be summarized as follows:
\vspace{-1mm}
\begin{itemize}
	\item We address the challenge of the FGVC by proposing a Grad-CAM guided channel-spatial attention module, which constrains the channel-spatial attention mechanism to focus on the parts that contribute most to the classification.
\vspace{-1mm}	
	\item We propose a Grad-CAM guided channel-spatial attention mechanism loss (GGAM-Loss) which employs the Grad-CAM to supervise and constrain the attention weights. Moreover, it is not limited to a specific network architecture.
\vspace{-1mm}	
	\item We conduct comprehensive experiments on the three commonly used FGVC datasets, \emph{i.e.}, CUB-$200$-$2011$~\cite{birds},  FGVC-Aircraft~\cite{air}， and Stanford Cars~\cite{cars} datasets. The results show the effectiveness of our method.
\end{itemize}
\vspace{-3mm}
\section{Methodology}

\begin{figure*}[!t]
    \centering 
    \includegraphics[width=1\linewidth]{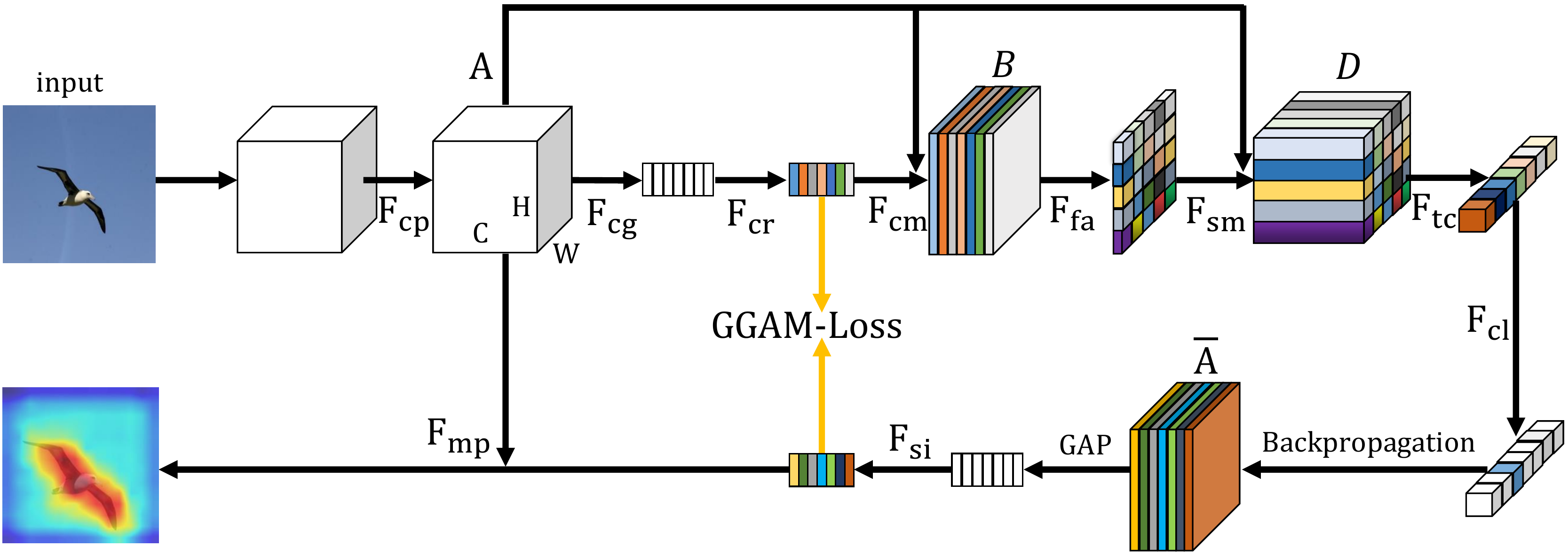}  
    \vspace{-8mm}
    \caption{The framework of our attention module. The upper line (from left to right) and the bottom line (from right to left) present the forward and the gradient backpropagation processes, respectively. A symmetrical Kullback-Leibler (KL) divergence between the weights of each channel in forward propagation and the weights of each feature map in the Grad-CAM is utilized as the loss function in backpropagation to supervise the channel-spatial attention.}\label{fig:structure}
\end{figure*}


\subsection{Channel-spatial Attention Mechanism}

The channel-spatial attention is a module that combines both the spatial attention and the channel attention. Specifically, as shown in Figure~\ref{fig:structure}, the input image is processed through a series of convolution and pooling operations $F_{cp}$ and feature maps denoted as $A=[a_{1}, a_{2},...,a_{C}]\in R^{C\times W\times H}$ are obtained, with height $H$, width $W$, and channel number $C$, respectively. Then we apply a global average pooling $F_{cg}$ to downsample each feature map in $A$ and a two-layer fully connected (FC) network $F_{cr}$ with softmax function to calculate the weights of each channel as the channel attention weights $S=[s_{1}, s_{2},...,s_{C}]\in R^{C}$, according to~\cite{SENet}. We rescale the original feature maps $A$ by $S$, which obtains the weighted feature maps $B=[b_{1},b_{1},...,b_{c}]\in R^{C\times W\times H}$ by $F_{cm}$ as

\begin{equation}
    b_c=F_{cm}(a_{c},s_{c})=a_{c}\cdot F_{cr}\left(F_{cg}(A)\right)_c,
\end{equation}

\noindent where $c=1,\cdots,C$.

After gaining the channel attention-weighted feature maps $B$, spatial attention is undertaken. Through the operation $F_{fa}$, which combines a channel-wise summation and a $2$D softmax function, the feature maps in $B$ are flattened along the channel dimension to obtain the spatial attention weights $T\in R^{W\times H}$. Then the channel-spatial attention-weighted feature maps $D=[d_{1},d_{1},...,d_{c}]\in R^{C\times W\times H}$ are obtained by rescaling $A$ with $T$ as 

\begin{equation}
    d_c=F_{sm}(a_{c},T)=a_{c} \odot F_{fa}(B),
\end{equation}

\noindent where $\odot$ is Hadamard product and
\begin{equation}
    T=F_{fa}(B)=\frac{\sum_{c=1}^C b_{c}}{\sum_{i=1}^W\sum_{j=1}^H \sum_{c=1}^C b_{c,i,j}}.
\end{equation}
Then the classification is undertaken according to $D$. $F_{tc}$ is the classifier with multiple FC layers and a softmax function.

\vspace{-2mm}
\subsection{Grad-CAM}

The Grad-CAM uses the class-specific gradient information and it flows into the final convolutional layer of a CNN to generate a heat map, which shows the main concentrated regions of the CNN. Specifically, as illustrated in Figure~\ref{fig:structure}, given an input image, we obtain the score $y^{k}$ for the predicted class $k$ before the last softmax function. Then, $y^{k}$ is propagated back to the elements of $A$ through the upper line and we gain the gradient $\frac{{\partial {y^k}}}{{\partial A_{c,i,j}}}$. The weights $\beta_c^k$ of the Grad-CAM, which represent the importance of feature map $c$ with the predicted class $k$, can be defined as

\begin{equation}
    \beta_c^k = \frac{1}{W\times H}\sum_{i=1}^W\sum_{j=1}^H\frac{{\partial {y^k}}}{{\partial A_{c,i,j}}}.
\end{equation}

\begin{table}[!t]
\vspace{-2.5mm}
    \caption{The statistics of the three FGVC datasets. \#Class, \#Training, and \#Test are class number, training sample number, and test sample number, respectively.}
    \centering
    \begin{tabular}{|c|c|c|c|}
        \hline
        Dataset & \#Class & \#Training & \#Test \\ 
        \hline
        \hline
        CUB-$200$-$2011$~\cite{birds} & $200$ & $5994$ & $5794$  \\ 
        \hline
        FGVC-Aircraft~\cite{air} & $100$ & $6667$ & $3333$ \\
        \hline
        Stanford Cars~\cite{cars} & $196$ & $8144$ & $8041$  \\
        \hline
    \end{tabular}
     \label{tab:datasets}
\end{table}
\begin{table*}[!t]
    
    \centering
    \caption{Classification accuracies (\%) on the CUB-$200$-$2011$, the FGVC-Aircraft, and the Stanford Cars datasets. The best results on each dataset are in~\textbf{bold}, and the second best results are in \underline{underline}.}

    \begin{tabular}{|c|c|c|c|c|}
        \hline
        Datasets& Base Model & CUB-$200$-$2011$ & FGVC-Aircraft & Stanford Cars \\ 
        \hline 
        \hline
        RA-CNN (CVPR$17$~\cite{RA-CNN}) & VGG$19$ & $85.30$ & $88.20$ &$92.50$ \\
        MA-CNN (ICCV$17$~\cite{MA-CNN}) & VGG$19$ &$84.92$ & $90.35$ &$92.80$ \\
        SENet (CVPR$18$~\cite{SENet}) & VGG$19$ & $84.75$ & $90.12$&$89.75$  \\ 
        SENet (CVPR$18$~\cite{SENet}) & ResNet$50$ & $86.78$ & $91.37$ &$93.10$ \\
        CBAM (ECCV$18$~\cite{CBAM})  & VGG$19$ & $84.92$ & $90.32$ &$91.12$ \\
        CBAM (ECCV$18$~\cite{CBAM})  & ResNet$50$ & $86.99$ & $91.91$ &$93.35$ \\
        DFL (CVPR$18$~\cite{DFL}) & ResNet$50$ & $87.40$ & $91.73$ &$93.11$ \\
        NTS (ECCV$18$~\cite{NTS}) & ResNet$50$ & $87.52$ & $91.48$ &$93.90$\\
        TASN(CVPR$2019$~\cite{TASN} )& VGG$19$ & $86.10$ & $90.83$ &$92.40$ \\
        TASN(CVPR$2019$~\cite{TASN})& ResNet$50$ & $87.90$ & $92.56$ &$93.80$ \\
        DCL(CVPR$2019$~\cite{DCL})& ResNet$50$ & $87.80$ & $\underline{93.00}$ &$\underline{94.50}$ \\
        ACNet(CVPR$2020$~\cite{ACNet})& ResNet$50$ & $\underline{88.10}$ & $92.40$ &$\boldsymbol{94.60}$ \\
        \hline
        Ours & VGG$19$ & $87.34$ & $91.55$ &$93.32$\\
        Ours & ResNet$50$ & $\boldsymbol{88.45}$ & $\boldsymbol{93.42}$ &$94.41$\\
        \hline
    \end{tabular}
    \label{tab:results}
\end{table*}
\vspace{-2mm}
\subsection{Grad-CAM guided channel-spatial attention loss}

In the FGVC, attention mechanism is introduced to ensure the CNN focus on more effective parts mainly, so as to improve the classification accuracy. As mentioned above, Grad-CAM can extract the key parts of the input image. 
In this section, we follow the same motivation and propose the Grad-CAM guided channel-spatial attention loss to enhance discriminative part searching and feature extraction abilities of CNNs in the FGVC. 

In Figure~\ref{fig:structure}, after the $F_{cr}$ operation, we can obtain the weights $S$ of each channel in $A$. Through the operation $F_{si}$, we apply a sigmoid function for $\beta_c^k,c=1,\cdots,C$, to scale their intervals and obtain $\tilde{\beta}^k=[\tilde{\beta}_{1}^k,\cdots,\tilde{\beta}_{C}^k]\in R^C$, where $\tilde{\beta}_{c}^k=\text{sigmoid}(\beta_c^k)$. As $\tilde{\beta}^k$ can reflect the contribution of each channel to the classification, we constrain the channel attention weights $S$ with it. Here, we propose the Grad-CAM guided channel-spatial attention mechanism loss, GGAM-Loss in short, to construct the regularization term. The GGAM-Loss ($L_{\text{GGAM}}$), which performs as a symmetrical Kullback-Leibler (KL) divergence between $S$ and $\tilde{\beta}^k$, can be defined as

\begin{equation}
    L_{\text{GGAM}} = \frac{1}{2}\left(\text{KL}(S||\tilde{\beta}^k) + \text{KL}(\tilde{\beta}^k||S)\right),
\end{equation}

\noindent where $\text{KL}(x||y)$ is the KL divergence from $x$ to $y$.

Moreover, as we use the original cross-entropy (CE) loss $L_{\text{CE}}$ for training the model as well, the total loss function $Loss$ of the whole network can be defined as

\begin{equation}
    Loss=L_{\text{CE}}+\lambda L_{\text{GGAM}},
\end{equation}

\noindent where $\lambda$ is a nonnegative multiplier.

\section{Experimental Results and Discussions}

\subsection{Datasets}

We evaluate our method on three challenging FGVC datasets, including CUB-$200$-$2011$~\cite{birds}, FGVC-Aircraft~\cite{air}, and~Stanford Cars~\cite{cars} datasets. The statistics of the datasets mentioned above, including class numbers and the training/testing sample numbers are shown in Table~\ref{tab:datasets}. We followed the same train/test splits as presented in the Table~\ref{tab:datasets}. For model training, we did not use artificially marked bounding box or part annotation.
\vspace{-1mm}
\subsection{Implementation Details}
\vspace{-0.35em}
In order to compare the proposed method with other attention mechanisms, we resized every image to $448\times448$, which is standard in the literatures~\cite{DCL,ACNet}. The backbones we used for extracting features were VGG$19$ and ResNet$50$ which were pre-trained on the ImageNet dataset. We used stochastic gradient descent optimizer. The weight dacay value and the momentum were kept as $1\times10^{-4}$ and $0.9$, respectively, with $100$ epochs. The learning rate of the FC layers was initially set at $0.1$ and we used the cosine anneal schedule update strategy~\cite{cos}. The learning rate of the pre-trained feature extaction layers was one-tenth of the FC layers. 
\begin{figure*}[!t]
    \centering 
    \includegraphics[width=1\linewidth]{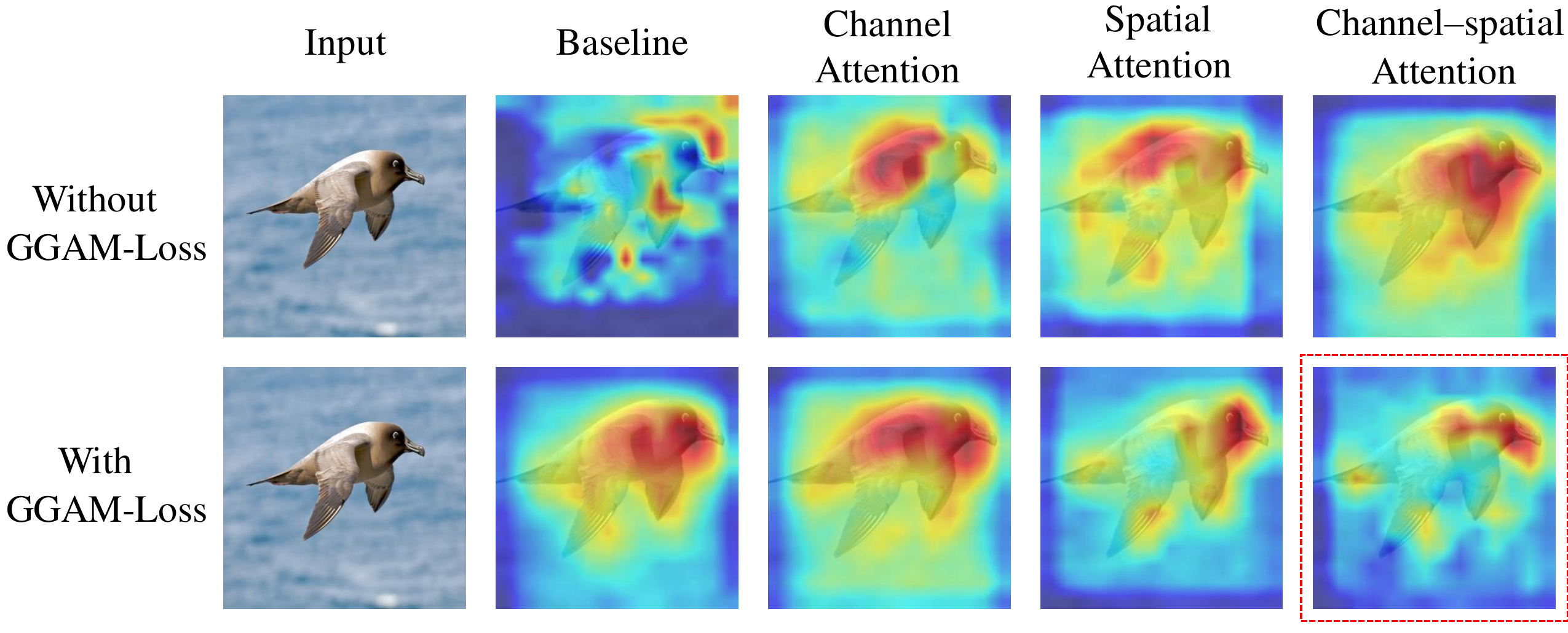}  
    \vspace{-6mm}
    \caption{Visualizations of the ablation models in Section~\ref{ssec:ablation}. The first column represents  the original image. The following four columns show visualization results of the baseline, the channel attention, the spatial attention, and the channel-spatial attention, respectively. The top row is trained without the GGAM-Loss, while the bottom row is trained with the GGAM-Loss.The red box indicates the visualization result of our proposed method.}\label{fig:visual}
\end{figure*}
\vspace{-1mm} 
\subsection{Experimental Results}

According to the aforementioned implementation details, the detailed results are listed in Table~\ref{tab:results}. Our method achieves significant performance improvement on all the three datasets and the evaluation results can be summarized as follows:

\begin{itemize}
    \item On the CUB-$200$-$2011$ dataset, our method achieves the best result on both VGG$19$ and ResNet$50$, respectively, comapred with their corresponding referred methods. Our method exceeds the second best method, TASN, by $1.24\%$ with the VGG$19$. In addition, compared with the leading result achieved by the ACNet, our method has improved the accuracy by $0.35\%$ with the ResNet$50$. 
    \vspace{-3mm}
    
    \item On the FGVC-Aircraft dataset, our method also obtains the best accuracy of $93.42\%$ with the ResNet$50$, around $0.4\%$ improvement than the DCL. With the VGG$19$, the result of our method also improves slightly.
    \vspace{-3mm}
    
    \item On the Stanford Cars dataset, our  method outperforms the most compared methods, especially with the same VGG$19$ backbone. The accuracy of the ACNet with the ResNet$50$ turns out $0.19\%$ better than ours. Note that the ACNet depends mainly on the addition of the network parameters and the complex training process to improve the accuracy, which is much more complex than ours.
\end{itemize}

\begin{table}[!t]

    \centering
    \vspace{-2.5mm}
    \caption{Ablation study of our method on classification accuracies (\%). Key modules of the proposed method, including the channel attention, the spatial attention, and the Grad-CAM are compared. 
    ``$\checkmark$'' represents the  module contained, otherwise ``$\times$''. The best result is in~\textbf{bold}. }
    \label{ablation}
    \resizebox{\linewidth}{!}{
    \begin{tabular}{|c|c|c|c|}
        \hline
        Spatial attention &Channel attention&GGAM-Loss& Accuracy  \\ 
        \hline 
        \hline
        $\times$&$\times$&$\times$&$85.10$  \\ 
        $\checkmark$ & $\times$&$\times$& $85.61$ \\
        $\times$ & $\checkmark$&$\times$& $85.39$ \\
        $\checkmark$ & $\checkmark$&$\times$& $86.86$ \\
        $\times$&$\times$&$\checkmark$&  $85.30$ \\
        $\checkmark$ & $\times$&$\checkmark$& $86.58$ \\
        $\times$ & $\checkmark$&$\checkmark$& $86.26$ \\
        $\checkmark$ &$\checkmark$&$\checkmark$&{$\boldsymbol{88.45}$} \\
        \hline
    \end{tabular}}
    
\end{table}

\subsection{Ablation Study}\label{ssec:ablation}

Attention mechanisms and Grad-CAM are major modules of our method, and the attention mechanisms include channel and spatial attention mechanisms. We analyze the influence of each module by the experimental results. The ablation experiments are all conducted on the CUB-$200$-$2011$ dataset and we use the ResNet$50$
as the base model if not particularly mentioned. The experimental results are shown in Table~\ref{ablation}.
\vspace{-1mm}
\begin{itemize}
    \item \textbf{Effectiveness of the attention mechanisms.} Compared with the baseline model, the spatial attention can improve performance by $0.51\%$ and the channel attention also has a slight promotion. In particular, the combination of channel and spatial attention obtains a $1.76\%$ increase on accuracy. This enhancement is obvious and shows that the channel-spatial attention is useful for the FGVC.
  
    \item \textbf{Effectiveness of the GGAM-Loss.} It can be seen that the classification accuracy of  each attention mechanism model is improved after adding the GGAM-Loss as the constraint for the attention mechanism. The above results demonstrate the effectiveness of the GGAM-Loss.
\end{itemize}

\subsection{Visualizations}

In order to better explain the improvement of our method, Figure~\ref{fig:visual} shows the visualizations of each model in Section~\ref{ssec:ablation}, which were generated by the Grad-CAM. The baseline cannot clearly focus on the right region of the object. With the addition of the attention mechanisms, the models tend to pay attention on the beak and the neck of the bird, which are discriminative parts. After adding the GGAM-Loss, the models can focus on more accurate discriminant characteristics and pay less attention to background information.

\subsection{Sensitivity Study of ${\lambda}$}

\begin{figure}[t]
 \vspace{-2mm}
    \centering 
    \includegraphics[width=0.9\linewidth]{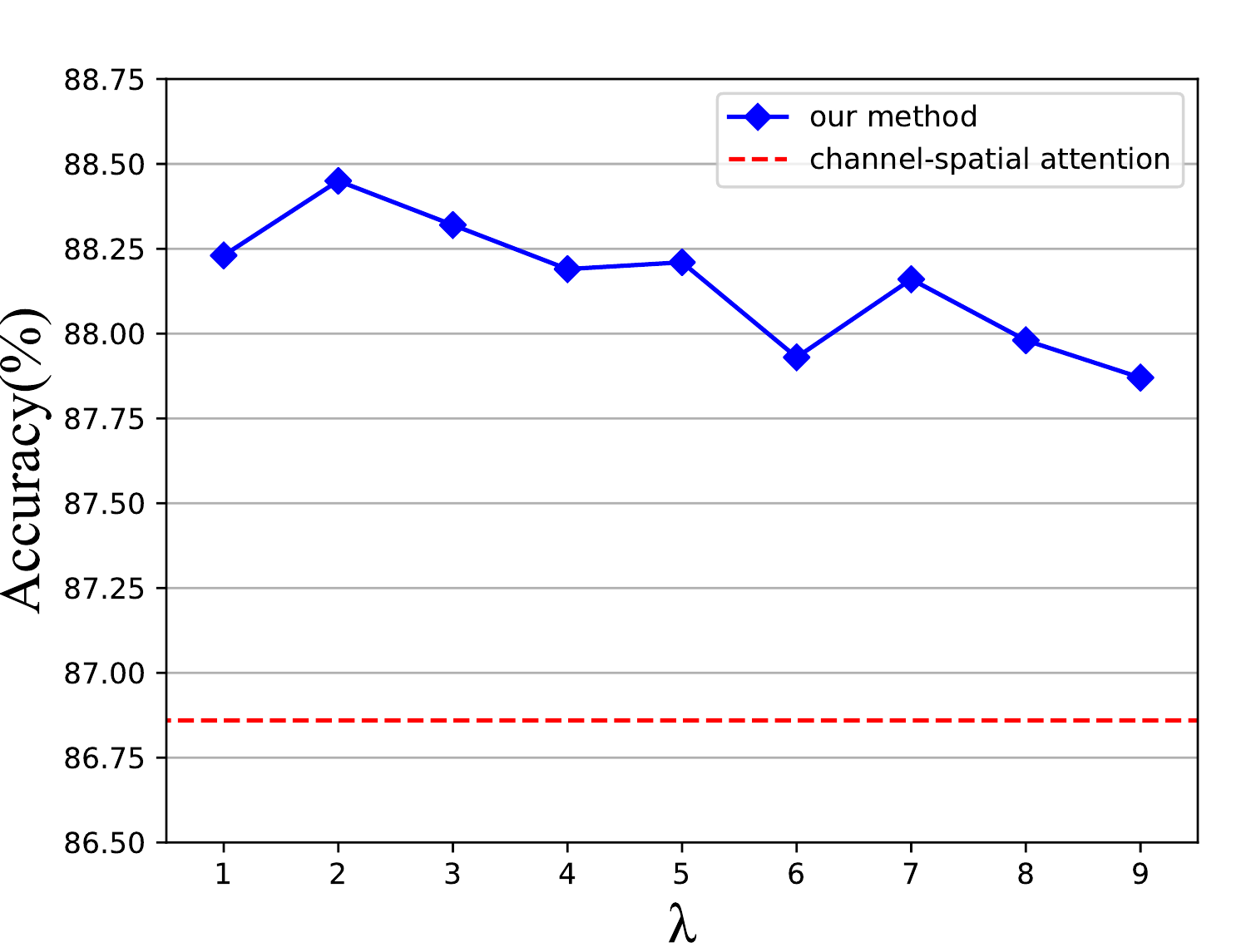}  
    \vspace{-4mm}
    \caption{Sensitivity study of $\lambda$ for our model on the CUB-$200$-$2011$ dataset.}\label{sensi}
\end{figure}

In order to evaluate the robustness of our method, we conduct the sensitivity study of the hyperparameter $\lambda$ to see whether the network performance changes a lot with a change  of $\lambda$. We conduct this study on the CUB-$200$-$2011$ dataset and we use the ResNet$50$ as the base model. We run the proposed model set with $\lambda$ varying from $1$ to $9$ with step size of $1$. The classification accuracies are shown in Figure~\ref{sensi}. From Figure~\ref{sensi}, it  can be observed that the performance of our method has always been better than the channel-spatial attention (without the GGAM-Loss) and does not change much by varying the value of $\lambda$, 
which proves the effectiveness and robustness of our method.

\section{Conclusions}
 
In this paper, we proposed a Grad-CAM guided channel-spatial attention mechanism loss (GGAM-Loss) for the FGVC task, which can constrain the channel-spatial attention module to focus on the most discriminative parts in the images. Note that the proposed GGAM-Loss can be also applied to other network architectures. The performance of our method is evaluated in the FGVC task and superior performance is achieved on three FGVC datasets (CUB-$200$-$2011$, Stanford Cars, and FGVC-Aircraft datasets). The effectiveness of the key modules of the proposed method were also evaluated. Visualizations of the feature maps illustrate the validity of the porposed method.

\bibliographystyle{IEEEbib}
\bibliography{icme2020template}

\begin{thebibliography}{10}

\bibitem{birds}
C.~Wah, S.~Branson, P.~Welinder, P.~Perona, and S.~Belongie,
\newblock ``The caltech-ucsd birds-200-2011 dataset,''
\newblock {\em California Institute of Technology}, 2011.

\bibitem{air}
S.~Maji, E.~Rahtu, J.~Kannala, M.~B. Blaschko, and A.~Vedaldi,
\newblock ``Fine-grained visual classification of aircraft,''
\newblock {\em CoRR}, vol. abs/1306.5151, 2013.

\bibitem{cars}
J.~Krause, M.~Stark, J.~Deng, and F.-F. Li,
\newblock ``3{D} object representations for fine-grained categorization,''
\newblock in {\em Proceedings of the International Conference on Computer
  Vision}, 2013.

\bibitem{flowers}
M.~{Nilsback} and A.~{Zisserman},
\newblock ``Automated flower classification over a large number of classes,''
\newblock in {\em Proceedings of the 2008 Sixth Indian Conference on Computer
  Vision, Graphics Image Processing}, 2008.

\bibitem{anno}
N.~Zhang, J.~Donahue, R.~Girshick, and T.~Darrell,
\newblock ``Part-based {R}-{CNNfs} for fine-grained category detection",''
\newblock in {\em Proceedings of the European Conference on Computer Vision},
  2014.

\bibitem{TASN}
H.~Zheng, J.~Fu, Z.~Zha, and J.~Luo,
\newblock ``Looking for the devil in the details: Learning trilinear attention
  sampling network for fine-grained image recognition,''
\newblock in {\em Proceedings of the Computer Vision and Pattern Recognition},
  2019.

\bibitem{NTS}
Z.~Yang, T.~Luo, D.~Wang, Z.~Hu, J.~Gao, and L.~Wang,
\newblock ``Learning to navigate for fine-grained classification,''
\newblock in {\em Proceedings of the European Conference on Computer Vision},
  2018.

\bibitem{BP}
C.~Yu, X.~Zhao, Q.~Zheng, P.~Zhang, and X.~You,
\newblock ``Hierarchical bilinear pooling for fine-grained visual
  recognition,''
\newblock in {\em Proceedings of the European Conference on Computer Vision},
  2018.

\bibitem{BT}
H.~Zheng, J.~Fu, Z.~Zha, and J.~Luo,
\newblock ``Learning deep bilinear transformation for fine-grained image
  representation,''
\newblock in {\em Advances in Neural Information Processing Systems}, 2019.

\bibitem{mamcloss}
M.~Sun, Y.~Yuan, F.~Zhou, and E.~Ding,
\newblock ``Multi-attention multi-class constraint for fine-grained image
  recognition,''
\newblock in {\em Proceedings of the European Conference on Computer Vision},
  2018.

\bibitem{mc}
D.~{Chang}, Y.~{Ding}, J.~{Xie}, A.~K. {Bhunia}, X.~{Li}, Z.~{Ma}, M.{Wu},
  J.~{Guo}, and Y.-Z. {Song},
\newblock ``The devil is in the channels: Mutual-channel loss for fine-grained
  image classification,''
\newblock {\em IEEE Transactions on Image Processing}, vol. 29, pp. 4683--4695,
  2020.

\bibitem{SENet}
J.~Hu, L.~Shen, and G.~Sun,
\newblock ``Squeeze-and-excitation networks,''
\newblock in {\em Proceedings of the Computer Vision and Pattern Recognition},
  2018.

\bibitem{MA-CNN}
H.~Zheng, J.~Fu, T.~Mei, and J.~Luo,
\newblock ``Learning multi-attention convolutional neural network for
  fine-grained image recognition,''
\newblock in {\em Proceedings of the International Conference on Computer
  Vision}, 2017.

\bibitem{RA-CNN}
J.~Fu, H.~Zheng, and T.~Mei,
\newblock ``Look closer to see better: Recurrent attention convolutional neural
  network for fine-grained image recognition,''
\newblock in {\em Proceedings of the Computer Vision and Pattern Recognition},
  2017.

\bibitem{PA-CNN}
H.~{Zheng}, J.~{Fu}, Z.{-}J. {Zha}, J.~{Luo}, and T.~{Mei},
\newblock ``Learning rich part hierarchies with progressive attention networks
  for fine-grained image recognition,''
\newblock {\em IEEE Transactions on Image Processing}, vol. 29, pp. 476--488,
  2020.

\bibitem{CBAM}
J.~Woo, S.and~Park, J.{-}Y. Lee, and I.~S. Kweon,
\newblock ``{CBAM}: Convolutional block attention module,''
\newblock in {\em Proceedings of the European Conference on Computer Vision},
  2018.

\bibitem{cam}
R.~R. Selvaraju, M.~Cogswell, A.~Das, R.~Vedantam, D.~Parikh, and D.~Batra,
\newblock ``{Grad-CAM}: Visual explanations from deep networks via
  gradient-based localization,''
\newblock in {\em Proceedings of the International Conference on Computer
  Vision}, 2017.

\bibitem{DFL}
Y.~Wang, V.~I. Morariu, and L.~S. Davis,
\newblock ``Learning a discriminative filter bank within a {CNN} for
  fine-grained recognition,''
\newblock in {\em Proceedings of the Computer Vision and Pattern Recognition},
  2018.

\bibitem{DCL}
Y.~Chen, Y.~Bai, W.~Zhang, and T.~Mei,
\newblock ``Destruction and construction learning for fine-grained image
  recognition,''
\newblock in {\em Proceedings of the Computer Vision and Pattern Recognition},
  2019.

\bibitem{ACNet}
R.~Ji, L.~Wen, L.~Zhang, D.~Du, Y.~Wu, C.~Zhao, X.~Liu, and F.~Huang,
\newblock ``Attention convolutional binary neural tree for fine-grained visual
  categorization,''
\newblock in {\em Proceedings of the Computer Vision and Pattern Recognition},
  2020.

\bibitem{cos}
G.~Huang, Y.~Li, G.~Pleiss, Z.~Liu, J.~E. Hopcroft, and K.~Q. Weinberger,
\newblock ``Snapshot ensembles: Train 1, get {M} for free,''
\newblock in {\em Proceedings of the International Conference on Learning
  Representations}, 2017.

\end{thebibliography}

\end{document}